\documentclass[10pt,journal,compsoc]{IEEEtran}

\ifCLASSOPTIONcompsoc
  \usepackage[nocompress]{cite}
\else
  \usepackage{cite}
\fi
\ifCLASSINFOpdf
\else
\fi
\usepackage{hyperref}       
\usepackage{url}            
\usepackage{booktabs}       
\usepackage{amsfonts}       
\usepackage{nicefrac}       
\usepackage{microtype}      

\usepackage{color,xcolor}
\usepackage{epsfig}
\usepackage{graphicx}
\usepackage{algorithm,algorithmic}

\usepackage{adjustbox}
\usepackage{array}
\usepackage{booktabs}
\usepackage{colortbl}
\usepackage{float,wrapfig}
\usepackage{framed}
\usepackage{hhline}
\usepackage{multirow}
\usepackage{tabularx}
\usepackage{subcaption} 
\usepackage[font=small]{caption}
\usepackage[percent]{overpic}

\usepackage{amsmath,amsfonts,amssymb}
\usepackage{amsthm} 
\usepackage{bm}
\usepackage{nicefrac}
\usepackage{microtype}
\usepackage{contour}
\usepackage{courier}

\usepackage{changepage}
\usepackage{extramarks}
\usepackage{fancyhdr}
\usepackage{lastpage}
\usepackage{setspace}
\usepackage{soul}
\usepackage{xspace}
\usepackage{cuted}
\usepackage{fancybox}
\usepackage{afterpage}
\usepackage{enumerate}
\usepackage{paralist}
\usepackage{enumitem} 
\usepackage{titlesec} 

\usepackage{url}
\usepackage{quoting}
\usepackage{epigraph}

\usepackage{comment}
\usepackage{pdfpages}

\newcolumntype{L}[1]{>{\raggedright\let\newline\\\arraybackslash\hspace{0pt}}m{#1}}
\newcolumntype{C}[1]{>{\centering\let\newline\\\arraybackslash\hspace{0pt}}m{#1}}
\newcolumntype{R}[1]{>{\raggedleft\let\newline\\\arraybackslash\hspace{0pt}}m{#1}}

\newcommand{\fig}[1]{Figure~\ref{#1}}
\newcommand{\tbl}[1]{Table~\ref{#1}}

\makeatletter
\DeclareRobustCommand\onedot{\futurelet\@let@token\@onedot}
\def\@onedot{\ifx\@let@token.\else.\null\fi\xspace}
\newcommand{\xhdr}[1]{\vspace{0pt} \noindent {\textbf{#1}}}

\def\eg{e.g\onedot}

\def\etc{etc\onedot} 
 
\def\etal{et al\onedot}
\makeatother

\definecolor{MyDarkBlue}{rgb}{0,0.08,1}
\definecolor{MyDarkGreen}{rgb}{0.02,0.6,0.02}
\definecolor{MyDarkRed}{rgb}{0.8,0.02,0.02}
\definecolor{MyDarkOrange}{rgb}{0.40,0.2,0.02}
\definecolor{MyPurple}{RGB}{111,0,255}
\definecolor{MyRed}{rgb}{1.0,0.0,0.0}
\definecolor{MyGold}{rgb}{0.75,0.6,0.12}
\definecolor{MyDarkgray}{rgb}{0.66, 0.66, 0.66}

\begin{document}
%
\title{Object-Centric Diagnosis of Visual Reasoning}
%
%
%
%

\author{Jianwei Yang\textsuperscript{1},
		Jiayuan Mao\textsuperscript{2},
		Jiajun Wu\textsuperscript{3},
		Devi Parikh\textsuperscript{1},\\
		David D. Cox\textsuperscript{4},
		Joshua B. Tenenbaum\textsuperscript{2}, 
		Chuang Gan\textsuperscript{4} \\
		\textsuperscript{1}{Georgia Institute of Technology},
		\textsuperscript{2}{MIT},
		\textsuperscript{3}{Stanford University},
		\textsuperscript{4}{MIT-IBM Watson AI Lab}\\
	}

%
%

\markboth{Journal of \LaTeX\ Class Files,~Vol.~14, No.~8, August~2015}%
{Shell \MakeLowercase{\textit{et al.}}: Bare Advanced Demo of IEEEtran.cls for IEEE Computer Society Journals}
%



\IEEEtitleabstractindextext{%
\begin{abstract}
When answering questions about an image, it not only needs \textit{knowing what} -- understanding the fine-grained contents (\eg, objects, relationships) in the image, but also \textit{telling why} -- reasoning over grounding visual cues to derive the answer for a question. Over the last few years, we have seen significant progresses on visual question answering. Though impressive as the accuracy grows, it still lags behind to get knowing whether these models are undertaking grounding visual reasoning or just leveraging spurious correlations in the training data. Recently, a number of works have attempted to answer this question from the perspectives such as grounding and robustness. However, most of them are either focusing on the language side, or coarsely studying the pixel-level attention maps. In this paper, by leveraging the step-wise object grounding annotations provided in GQA dataset, we first present systematical \textit{object-centric} diagnosis of visual reasoning on grounding and robustness, particularly on the vision side. According to the extensive comparisons across different models, we find that even models with high accuracy are not good at grounding objects precisely, nor robust to visual content perturbations. In contrast, symbolic and modular models have relatively better grounding and robustness, though at the cost of accuracy. To reconcile these different aspects, we further develop a diagnostic model, namely Graph Reasoning Machine, which performs neuro-symbolic reasoning over scene graphs on realistic images. Our model replaces purely symbolic visual representation with probabilistic scene graph and then applies teacher-forcing training for the visual reasoning module. The designed model improves the performance on all three metrics over the vanilla neural-symbolic model while inheriting the transparency. Further ablation studies suggest that this improvement is mainly due to more accurate image understanding and proper intermediate reasoning supervisions.

\end{abstract}

}

\maketitle

\IEEEdisplaynontitleabstractindextext

%
\IEEEpeerreviewmaketitle

\begin{figure}[!ht]
    \centering
  	\includegraphics[width=\linewidth]{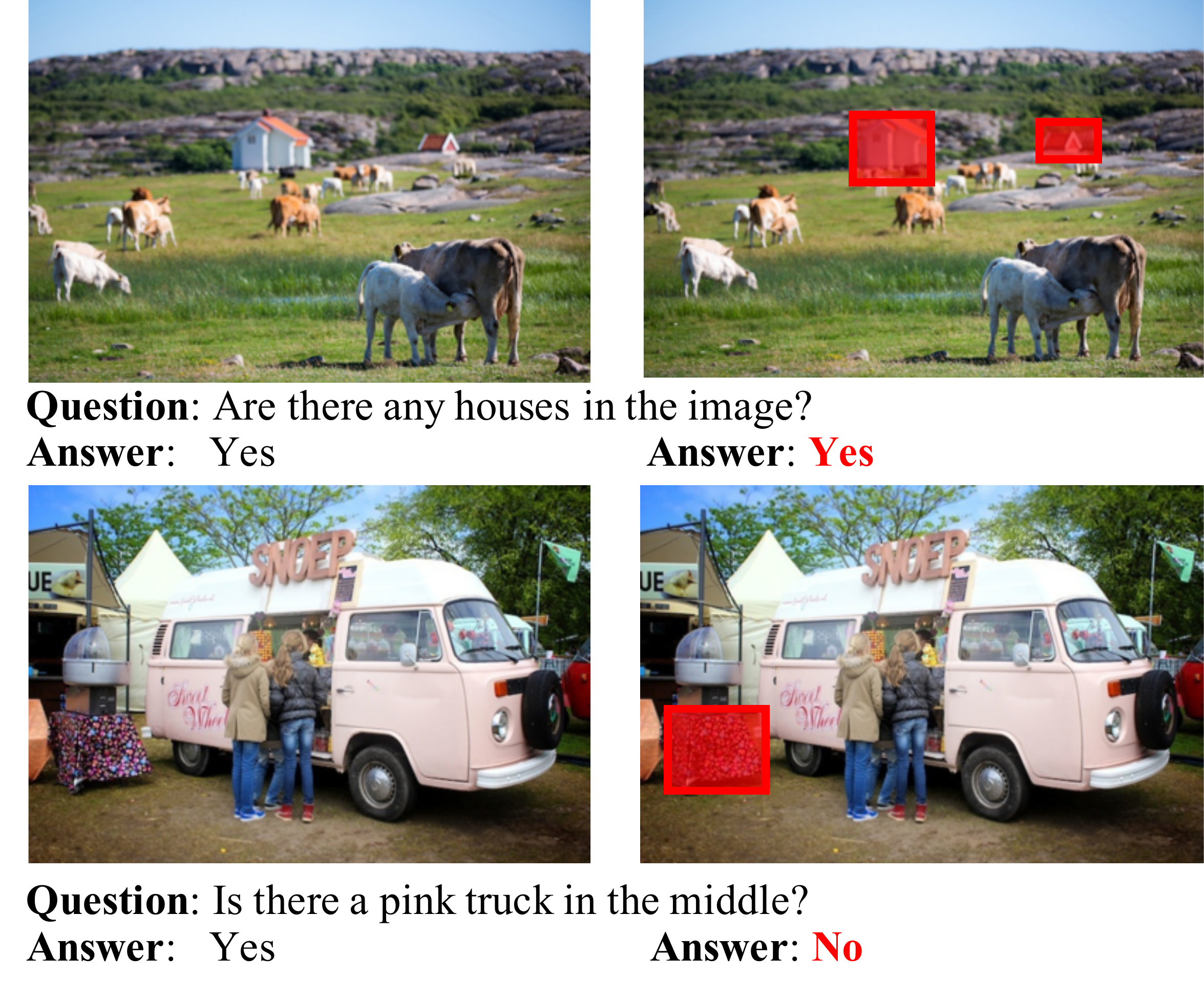}
    \caption{Two simple probing tests show that current VQA system fails to grounding on the correct objects and robust to perturbations: at top row, model gives the same answer after removing foreground objects ``houses''; At bottom row, however, covering a background object inverts the answer.}
    \label{Fig:Teaser}%
\end{figure}

\IEEEraisesectionheading{\section{Introduction}}
\IEEEPARstart{B}uilding AI systems that can understand and reason over vision and language data has been a long-standing challenge in the community\cite{antol2015vqa,johnson2017clevr,agrawal2018don,hudson2019gqa}. In this paper, we study the problem of visual question answering (VQA), a task to predict the answer to a question, taking an image as the context. VQA has been attributed as a representative challenge at the intersection of vision and language since it requires a sophisticated reasoning over the visual and language contents. Current VQA systems still make some ridiculous mistakes which humans can easily resolve. 
As shown in Fig.~\ref{Fig:Teaser} top row, a carefully designed VQA model\footnote{we use the model provided by Pythia~\cite{singh2018pythia}} thinks houses still there in the image even they are already occluded by the red boxes. Similarly, when being asked ``is there a pink truck in the middle?" in Fig.~\ref{Fig:Teaser} bottom row, the model gives opposite answers before and after we cover a background object in the image. These two examples indicate that there is still a gap between current AI systems and humans in terms of visual reasoning, which requires grounding on image contents and robustness to background perturbations. 

Recently, tremendous progresses have been made on VQA by building large and diagnostic datasets \cite{antol2015vqa,johnson2017clevr,agrawal2018don,hudson2019gqa,su2019vl,yi2019clevrer} and designing end-to-end or modular models \cite{antol2015vqa, lu2016hierarchical,zhang2016yin,agrawal2018don,anderson2018bottom, nsvqa, hu2019language}. 
Though the accuracy has been boosted drastically in the last few years~\cite{anderson2018bottom,hudson2018compositional,hudson2019learning}, it still lags behind to inspect whether those models truly understand scenes and texts and perform visual reasoning, or barely overfit to spurious correlations in the training data. Recently, some efforts have been made to study these aspects including a) \textbf{Grounding} by visualizing the activation maps on input images~\cite{selvaraju2017gradcam, patro2019u} and further aligning their activation maps to human attentions~\cite{das2017human,selvaraju2019taking}; b) \textbf{Robustness}\footnote{This is also called consistency in some literature.} by perturbing questions with commonsense-based or logic-based consistency~\cite{mudrakarta2018did,ray2018make,ray2019sunny}, or imposing different question and answer distributions during training and testing~\cite{agrawal2018don,ramakrishnan2018overcoming}. However, for grounding, activation maps are far precise to depict whether the reasoning is on or off a specific object, not to say the difficulty in quantification and generalization to modular networks or symbolic models~\cite{hudson2019gqa}. For robustness, most of the works focus on the language side by probing the system with adversarial or counterfactual questions~\cite{cadene2019rubi,clark2019don,niu2020counterfactual}. In this paper, we focus on the vision side and propose object-centric diagnosis methods to inspect the visual reasoning capacity for various types of models. The reason to perform object-centric analysis is two-fold: 1) It aligns better with human intuition in that humans can easily understand whether a VQA model grounds itself on a specific object by looking at the bounding boxes; 2) It makes the quantitative measurement simpler and can leverage the rich object annotations in existing dataset, such as GQA~\cite{hudson2019gqa}. Based on these annotations, we exploit quantitative metrics to measure
\begin{itemize}
    \item Grounding -- whether a VQA model is attending or pointing to the target objects in the image. For example, the houses should be the target, rather than other objects in Fig.~\ref{Fig:Teaser} top row. The answer should therefore be ``No'' if they are occluded.
    \item Robustness -- whether a VQA model is robust to probings such as removing a background object. In Fig.~\ref{Fig:Teaser} bottom row, the model is not robust enough since it outputs incoherent answers when we occlude the irrelevant object.
\end{itemize}

Using these two metrics, we conduct extensive analysis on various VQA models to get a holistic sense about their visual reasoning ability. 
In our comparisons, we observe that the models with high accuracy do not necessarily have a better grounding or robustness, while modular models such as Meta-Modular Network~\cite{chen2019meta} are the other way around. To better understand the effects of different components in the models, we further develop a diagnostic model, namely Graph Reasoning Machine. It is inspired by the neural-symbolic VQA model (NS-VQA)~\cite{nsvqa}, which was originally built for visual reasoning on CLEVR~\cite{johnson2017clevr}. Likewise, our model consists of three modules: 1) scene graph generator; 2) program generator and 3) program executor. However, it differs from NS-VQA from three aspects: 1) we apply a scene graph generator on top of the realistic images to extract the scene graphs with a much larger vocabulary size; 2) instead of using a pure symbolic scene graph representation, we extract a probabilistic scene graph and 3) our model has a learnable program executor so that it can learn how to execute on the probabilistic scene graph during training. Based on these novel designs, we achieve better performance at all three aspects including accuracy, grounding and robustness. Due to its transparency, we further delve into the model to understand the reasons for successes and failures.

To summarize, our main contributions are three-fold. First, we propose object-centric diagnosis of visual reasoning for VQA models. With respect to the visual contents, we investigate how these models perform with respect to two quantitative metrics, grounding and robustness. Second, we conduct extensive comparisons to analyze the visual reasoning capacity for various models by leveraging the fine-grained object grounding annotations in GQA dataset. Third, we further develop a diagnostic model to perform in-depth analysis on different components to narrow down the reasons for successes and failures in visual question answering.

\section{Related Work}
\begin{table*}[!ht]
\small
\centering
\begin{tabular}{l c c c c}
\toprule
Method & Visual Representation & Question Representation & Viusal Reasoning \\
\midrule
Bottom-Up~\cite{anderson2018bottom} & Object & Recurrent Neural Network & Sequential attention  \\
LCGN~\cite{hu2019language} & Object & Recurrent Neural Network  & Sequential attention \\
NSM~\cite{hudson2019learning} & Scene graph & Recurrent Neural Network & Sequential attention  \\
LXMERT~\cite{tan2019lxmert} & Object & Self-attention & Self-attention \\
NS-VQA~\cite{nsvqa} & Scene graph & Symbolic program & Sequential execution \\
MMN~\cite{chen2019meta} & Object & Neural program & Sequential execution  \\
\bottomrule
\end{tabular}
\caption{An overview of VQA models on GQA by comparing their visual representations, question representations and reasoning method.}
\label{Table:ReasoningComparison}
\end{table*}

\subsection{Visual Question Answering} 
Visual Question Answering (VQA) has received considerable attentions in recent years~\cite{antol2015vqa,lu2016hierarchical,zhu2016visual7w,gan2017vqs,agrawal2018don,goyal2017making,hu2017learning,trott2017interpretable,anderson2018bottom,singh2018pythia,hu2019language,andreas2016neural}. To address this task, the model is usually comprised of three components: 1) a visual encoder to extract image representation; 2) a language encoder to encode the question, and 3) a multi-modal fusion module to derive the answer. Earlier works such as~\cite{yang2016stacked,lu2016hierarchical} used a regular CNN to extract the grid of image features and then a RNN to encode the question. Since the development of Bottom-Up and Top-Down (BUTD) model~\cite{anderson2018bottom}, object-centric features have become the most commonly used visual representations~\cite{singh2018pythia,hu2019language}. At most recent, inspired by the success of BERT pretraining~\cite{devlin2018bert}, a number of visual-linguistic pretraining pipelines are proposed to learn a generic cross-modality representation which proves to be effective on VQA task~\cite{Lu2019,tan2019lxmert,Zhou2019a,li2020oscar}. Though impressive results have been achieved, it remains unclear whether these end-to-end models reason over scenes and language or just capture spurious correlations in the data. To demystify this, a number of diagnostic datasets have been proposed. In~\cite{johnson2017clevr}, Justin \etal introduced a synthetic image dataset to diagnose the visual reasoning capacity. Hudson and Manning~\cite{hudson2019gqa} further extended it to realistic images and build a new dataset GQA. In this dataset, the authors provide fine-grained scene graphs and also the question programs as the extra annotations. Recently, we have seen a number of work on this dataset~\cite{hudson2019gqa, hudson2018compositional, hudson2019learning, chen2019meta, amizadeh2020neuro}. However, none of them have utilized the fine-grained annotations to perform systematical analysis on the visual reasoning capacity.

\subsection{Grounding} 
The notion of visual grounding has been discussed broadly not only on visual question answering~\cite{lu2016hierarchical,zhu2016visual7w,gan2017vqs,agrawal2018don,goyal2017making,hu2017learning,trott2017interpretable,anderson2018bottom,singh2018pythia,hu2019language,andreas2016neural,chen2019meta,nsvqa,mao2019neuro,selvaraju2017gradcam}, but also in other vision-language tasks such as image captioning~\cite{lu2017knowing,Lu_2018_CVPR,anderson2018bottom,yang2018graph,fan2019bridging}, image co-reference resolution~\cite{plummer2015flickr30k,kong2014you,ramanathan2014linking,yu2016modeling,yu2018mattnet,liu2019clevr}, etc. However, most of them leveraged the intuition of grounding to guide the model design or dataset collection. A few works~\cite{selvaraju2017gradcam,patro2019u} have proposed to interpret the grounding based on the class activation maps (CAM). By further using the limited human attention annotations~\cite{das2017human}, the grounding performance can be improved by designing better models~\cite{selvaraju2020squinting} or regularization~\cite{shrestha2020negative}. On one hand, these interpretation methods diverge from the object-centric intuition of humans when answering a visual question. On the other hand, it is difficult to quantitatively measure how much grounding a VQA model is. Moreover, since it is based on gradient back-propagation, it is hard to apply it to a broad types of reasoning models, such as NS-VQA~\cite{nsvqa}. In GQA~\cite{hudson2019gqa}, the authors also provided a method to measure grounding. However, it is only applied to their own model, and not able to generalize to other models.

\subsection{Robustness} 
In VQA, robustness means the models should maintain coherence of answers to the variety of images and questions. Recently, many previous works focus on the language side and argue that current VQA models answer questions by heavily leveraging the language priors~\cite{goyal2017making, agrawal2018don,ramakrishnan2018overcoming}. To address this, they resort to collect new datasets~\cite{goyal2017making,agrawal2018don,huang2019assessing,huang2019novel,mudrakarta2018did,ray2018make,ray2019sunny} or design better model~\cite{agrawal2018don,ramakrishnan2018overcoming,ray2019sunny,niu2020counterfactual}. Unlike the language side, robustness has been less touched on the image side. Recently, there are some concurrent works that modify the image contents to study and improve the robustness of VQA models. In~\cite{patro2020robust}, the authors proposed a correlated module to improve the robustness of VQA models to the pixel-level adversarial attacks. In contrast, Agarwal~\etal instead focus on semantic objects by erasing them in the image to improve the robustness of VQA models~\cite{Agarwal2019TowardsCV}. Similarly in~\cite{chen2020counterfactual,gokhale2020mutant}, a counterfactual sample synthesizing strategy is proposed by masking critical objects in the original image. All of these work use VQA~\cite{antol2015vqa}, which is lack of fine-grained grounding annotations. Therefore, the critical objects can only be roughly predicted using word-region matching~\cite{Agarwal2019TowardsCV,gokhale2020mutant} or object-level Grad-CAM~\cite{chen2020counterfactual}. In this paper, however, we study the robustness of various VQA models on GQA dataset. Leveraging the fine-grained annotations, we present a quantitative measurement for robustness and provide a comprehensive comparison on various VQA models.

\section{Object-Centric Diagnosis}
In this section, we introduce the object-centric diagnosis method on grounding and robustness. We first summarize the generic pipeline used in current VQA models, including both end-to-end and neural-symbolic models. Then, we elaborate on how we develop the measurements for all these VQA models.

\subsection{Preliminaries}
\label{Sec:Related}
In a VQA model, an image $\bm{I}$ and a question $\bm{Q}$ are given and the goal is to predict most likely answer $\bm{a}^*$ to the question:
\begin{equation}
\small
\bm{a}^* = \arg\max_{\bm{a}} p(\bm{a}|\bm{I}, \bm{Q})
\end{equation}

Earlier models usually learn to encode images and questions into hidden vectors and predict the answer through a classifier~\cite{zhou2015simple,antol2015vqa,Jabri2016Revisiting,lu2016hierarchical,hu2017learning,hudson2018compositional}. However, since the significant improvements shown in \cite{teney2017graph,anderson2018bottom}, recent models introduce the notion of object and factorize the process into three components:
\begin{equation}
    \bm{a}^* = \arg\max_{\bm{a}} p(\bm{a}|\bm{G}, \bm{R})p(\bm{G}|\bm{I}) p(\bm{R}|\bm{Q})
    \label{Eq:VQA_Factorized}
\end{equation}
where $p(\bm{G}|\bm{I})$ is responsible to extract the object-centric representation from the image, $p(\bm{R}|\textbf{Q})$ is used to extract the question representation while $p(\bm{a}|\bm{G},\bm{R})$ for deriving the answer given the intermediate visual and question representations. To represent an image, we can extract the objects \cite{ren2015faster} solely or augment it with a scene graph~\cite{johnson2015image,yang2018graph,zellers2018neural}. Moreover, these object-centric representations can be either embeddings~\cite{anderson2018bottom, hu2019language,hudson2019learning,tan2019lxmert,chen2019meta} or semantic symbols~\cite{nsvqa}. There are also various ways to represent the questions, such as recurrent neural networks~\cite{anderson2018bottom,hu2019language,hudson2019learning}, self-attention layers~\cite{tan2019lxmert} or programs~\cite{nsvqa,chen2019meta}. Depending on how the image and question are represented, $p(\bm{a}|\bm{G},\bm{R})$ can be instantiated by a sequential attention process~\cite{anderson2018bottom,hu2019language,hudson2019learning}, self-attention~\cite{tan2019lxmert} or program execution~\cite{nsvqa,chen2019meta}. A full comparison with respect to these three components is shown in \tbl{Table:ReasoningComparison}. The commonly used object-centric representation and attention mechanism make it possible to extract the grounded objects (a formal definition will be given in Sec.~\ref{Sec:Measurements}) for different models, which offers us the opportunity to perform object-centric diagnosis for all of them. In the following, we will elaborate the measurement of grounding and robustness.

\begin{table*}[!ht]
\setlength{\tabcolsep}{4.3pt}
\centering
\begin{tabular}{l c c c c c c c}
\toprule
Model & Bottom-Up~\cite{anderson2018bottom} & LCGN~\cite{hu2019language} & NSM~\cite{hudson2019learning} & LXMERT~\cite{tan2019lxmert} & NS-VQA~\cite{nsvqa} & MMN~\cite{chen2019meta} \\
\midrule
Size (MB) & 19.63 & 10.63 & n/a & \textbf{202.13} & 12.81 & 36.27 \\
Accuracy & 49.74 & 56.10 & \textbf{63.17} & 60.30 & 28.81 & 59.33 \\
\bottomrule
\end{tabular}
\vspace{2pt}
\caption{Mode size and accuracy on the GQA test split. We cannot get model size for NSM~\cite{hudson2019learning} as it is not open-sourced.}
\label{Table:AccuracyComparison}
\end{table*}
\subsection{Measurements}
\label{Sec:Measurements}
In this part, we present a new and generic method to measure how well a VQA model performs regarding grounding and robustness. Without the loss of generality, visual reasoning can be formulated as $T$ soft/hard attention/selection steps on the detected objects. For clarity, we make the following definitions:
\setdefaultleftmargin{1em}{1em}{}{}{}{}
\begin{compactitem}
\item \textbf{Detected objects}. Detected objects $\bm{O}_d=\{\bm{o}^{1}_d, ...,\bm{o}^{K_d}_d\}$ is the set of $K_d$ objects localized using an object detector or scene graph generator. Each object is represented by its category and bounding box location. As shown in \fig{Fig:Teaser}(a), the detected objects include `house' and many other objects such `grass', `cow', \etc.
\item \textbf{Referred objects}. At reasoning step $t$, we define the referred objects as those the current reasoning step cares about. For example in \fig{Fig:Teaser}(b), a question ``how many people are standing beside the truck?'' has an intermediate reasoning step to localize the truck. At this reasoning step, the referred objects should be the truck in the middle. We denote the $K^t_r$ referred objects at step $t$ by $\bm{O}_r^t = \{\bm{o}^{t,1}_r, ...,\bm{o}^{t,K^t_r}_r\}$. 

\item \textbf{Grounded objects}. We define the grounded objects $\bm{O}_g$ as the final referred objects the visual reasoning should attend to, which is thus equivalent to $\bm{O}^T_r$. Take the same example as above, the two persons in \fig{Fig:Teaser}(b) are the grounded objects based on which the answer will be given.

\item \textbf{Foreground objects}. $\bm{O}_f=\{\bm{o}^{1}_f, ...,\bm{o}^{K_f}_f\}$ is a subset of detected objects $\bm{O}_d$, which have overlaps with the combination of referred objects $\bm{O}_r^t$ at all steps and grounded objects $\bm{O}_g$. Here, the overlap is measured by computing the intersection over union (IoU) over bounding boxes.
\item \textbf{Background objects}. $\bm{O}_b=\{\bm{o}^{1}_b, ...,\bm{o}^{K_b}_b\}$ is a subset of detected objects $\bm{O}_d$ which, however, do not have any overlaps with the referred objects $\bm{O}_r^t$ at any steps.
\end{compactitem}

For the ground-truth referred and grounded objects, we can obtain them using the ground-truth scene graph and question program annotations in GQA dataset. However, the extraction of referred and grounded objects for different models depends on the specific attention mechanism. Below, we introduce the proposed method to find the referred and grounded objects for different models.

For models that use sequential attention such as Bottom-Up~\cite{anderson2018bottom} and LCGN~\cite{hu2019language}, there are one or multiple attention steps between the question and detected objects. However, since there is no explicit reasoning in the intermediate steps, we can hardly obtain the referred objects precisely. As such, we extract the grounded objects based on the attention scores at the last step. For self-attention based model LXMERT~\cite{tan2019lxmert}, the attention is performed layer by layer with multi-head. Specifically, LXMERT has 5 cross-modality self-attention layers, and each layer has 12 attention heads. As such, we can obtain the cross-modality attention score matrix $S \in \mathcal{R}^{5 \times 12 \times L \times K_d}$, where $L$ is the length of the question, and $K_d$ is the number of objects. To aggregate attention scores over all $K_d$ object, we take the average over all self-attention layers, after taking the maximum over all heads to obtain the attention scores between question tokens and objects, $\bm{s} \in \mathcal{R}^{L \times K_d}$. The final group of methods, such as NS-VQA~\cite{nsvqa} and MMN~\cite{chen2019meta}, explicitly use a program executor to execute the question programs. The advantage of these models is that we can interpret their intermediate reasoning steps. For these methods, we can extract the referred objects and corresponding scores at each step. However, for comparison with other work, we only consider the grounded objects at last step. Below we explain how to measure the grounding and robustness.

\xhdr{Grounding}. Ideally, a visual reasoning model should attend exactly to the referred and grounded objects at all reasoning steps. Since there are discrepancies between the detected objects and ground-truth objects, we propose to compute the average precision (AP) between grounded objects detected by the model and ground-truth objects. Specifically, at final reasoning step, we get attention score over the detected objects $\bm{O}_d$ for the model. Then, we compare these detected objects with the ground-truth grounded objects $\bm{O}_g^t$, by setting an IoU threshold (0.5 in our paper). Finally, we vary the threshold for the attention score and obtain the recall and precision at different points, which are then averaged across different recall checkpoints. Based on this, we can compute the grounding score for different models. Since not all models have explicit intermediate reasoning steps, we report the grounding scores at the final reasoning step for comparisons.

\xhdr{Robustness}. As illustrated in Fig.~\ref{Fig:Teaser}(b), when removing irrelevant objects from images, we expect models to still hold the correct answer. Motivated by this, we remove the irrelevant objects and then measure how the answers are affected. Ideally, if a VQA model is perfect, the answers will be all correct and not changed after this erase since all referred and grounded objects are reserved. In practice, models learn spurious correlations from the data more or less. Therefore, the flipping rates are no longer zero and thus can be used as an indicator of model's robustness to image perturbations. To measure the robustness, we regard all background objects $\bm{O}_b$ as irrelevant objects, and then count the answer flips after removing them from the inputs of visual reasoning module. Specifically, the answer flipping rates are measured in two different ways: 1) from a correct answer to an incorrect one (C$\rightarrow$I) and 2) from an incorrect to a correct answer (I$\rightarrow$C). Note that we compare the answer flipping rate instead of accuracy, because accuracy is prone to be biased for several reasons: 1) the inputs to visual reasoning module are different; 2) different models have distinct capacities and 3) they leverage language priors to different extents. Besides the background removal, we also report the flipping rates for foreground removal and randomized visual representations.

\subsection{Experiments}

\xhdr{Datasets}. As mentioned above, we use GQA~\cite{hudson2019gqa} in our experiments considering it provides fine-grained scene graph annotations and reasoning programs. Specifically, we use the balanced set which contains 72,140, 10,233, and 398 images for training, validation, and testdev, respectively. Each image in training and validation set has a ground-truth scene graph and around 13 questions associated with step-wise question programs. In the testdev set, no scene graph is provided and questions have a similar distribution to the test set. We use the training split to train models, and report performance on the validation and testdev sets.

\xhdr{Models.} We perform our analyses on the models listed in Table~\ref{Table:ReasoningComparison}, except for NSM~\cite{hudson2019learning} because it does not have an open-sourced implementation. These five models we choose cover the variety of visual reasoning modules on GQA. For NS-VQA, since it was originally designed for CLEVR~\cite{johnson2017clevr}, we train a new scene graph generator~\cite{yang2018graph} on the GQA training set as the new vision backbone to detect objects, attributes, and relationships.

\begin{table*}[t]
\centering
\begin{tabular}{l c c c c c c}
\toprule
\multirow{2}{*}{Model} & \multicolumn{2}{c}{Question Type} & \multicolumn{3}{c}{Semantic Type} & \multirow{2}{*}{All}  \\
\cmidrule(lr){2-3}\cmidrule(lr){4-6}
& Open & Binary & Object & Attribute & Relation & \\
\midrule
Bottom-Up~\cite{anderson2018bottom}    & 38.30    & 32.45     & 32.10   & 31.39 & 35.71 & 36.24 \\
LCGN~\cite{hu2019language}       & 36.71    & 19.63    & 19.16  & 31.02  & 31.91 & 33.00  \\
LXMERT~\cite{tan2019lxmert} & 26.62    & 12.53     & 27.39     & 16.32      & 22.66 & 21.31  \\
NS-VQA~\cite{nsvqa}     & 25.25     & 36.78     & 29.59     & 44.70     & 20.99 & 29.57  \\
MMN~\cite{chen2019meta} & \textbf{52.75}     & \textbf{50.49}   & \textbf{49.64}  & \textbf{57.82}  & \textbf{46.51}  & \textbf{52.26}   \\
\bottomrule
\end{tabular}
\caption{Object grounding scores of different models evaluated on the GQA validation set.}
\label{Table:Grounding}
\end{table*}

\begin{table*}[t]
\setlength{\tabcolsep}{3.5pt}
\centering
\begin{tabular}{l c c c c c c c c c c c c c c}
\toprule
\multirow{2}{*}{Method} & Val & \multicolumn{3}{c}{BG. Removal} & & \multicolumn{3}{c}{FG. Removal} & & \multicolumn{3}{c}{Randomized Obj.}  \\
\cmidrule{3-5}\cmidrule{7-9}\cmidrule{11-13}
& Acc. & Acc.  &C$\rightarrow$I & I$\rightarrow$C & & Acc. & C$\rightarrow$I & I$\rightarrow$C & & Acc. & C$\rightarrow$I & I$\rightarrow$C \\
\midrule
Buttom-Up~\cite{anderson2018bottom}    & 53.73  & 56.30  & 5.92     & 12.71 &  & 46.99 & 42.10 & 7.65 & & 32.81 & 64.17 & 21.92 \\
LCGN~\cite{hu2019language}       &  63.87  & 65.79   & 4.82  & 10.11 &  & 44.94 & 33.74 & 7.14 & & 37.71 & 51.81 & 19.15     \\
NS-VQA~\cite{nsvqa}      & 40.49     & 38.64  & 5.77  & 1.23     & & 25.41 & 59.38 &1.10  & & 6.99 & 85.38 & 0.00  \\
MMN~\cite{chen2019meta} & {68.87} & {70.05}    & 6.27     & \textbf{17.27}  &  & 45.56 & 38.36 & 9.92 & & 37.12 & 53.49 & 16.37 \\
\bottomrule
\end{tabular}
\vspace{5pt}
\caption{Model accuracy and flipping rates after different type of image perturbations. The C$\rightarrow$I denotes the percentage of questions that were answered correctly before the modification but wrong afterward, while I$\rightarrow$C denotes the reversed direction. We perform three types of modifications: background removal, foreground removal and object randomization.}
\label{Table:Robustness}
\end{table*}

\begin{figure*}[t]
    \centering
    \includegraphics[width=1\linewidth]{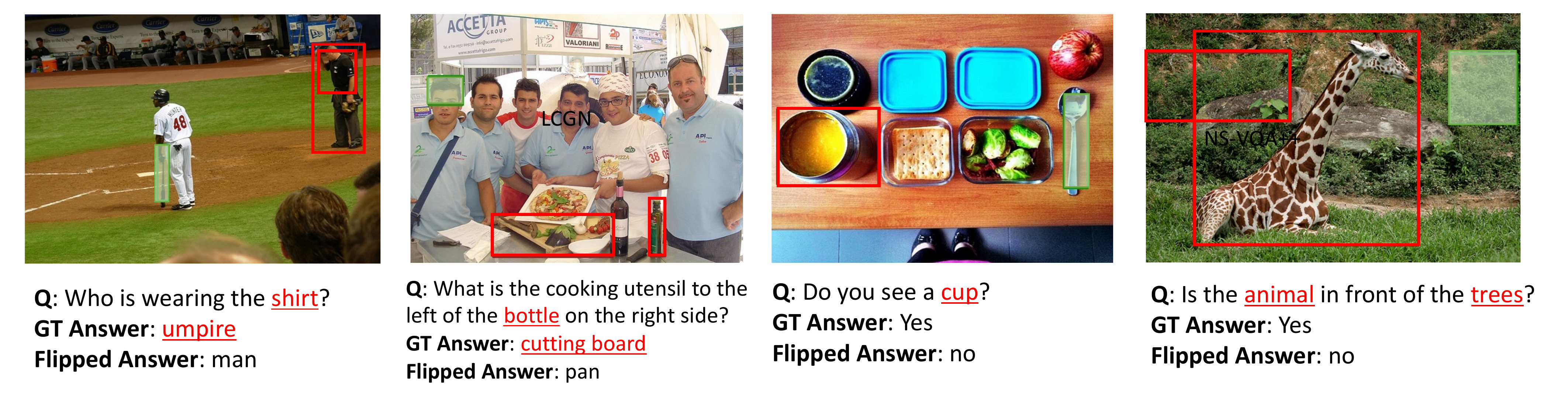}
    \vspace{-18pt}
    \caption{Examples showing answers flipped by removing a single irrelevant objects from images. From left to right, the methods are Bottom-Up~\cite{anderson2018bottom}, LCGN~\cite{hu2019language}, MMN~\cite{chen2019meta}, and NS-VQA~\cite{nsvqa}. The red boxes show ground-truth grounded objects for the question below, while the green box is the one we remove from the image.}
    \label{Fig:Robustness}
\end{figure*}

\begin{figure*}[!ht]
    \centering
    \includegraphics[width=0.8\linewidth]{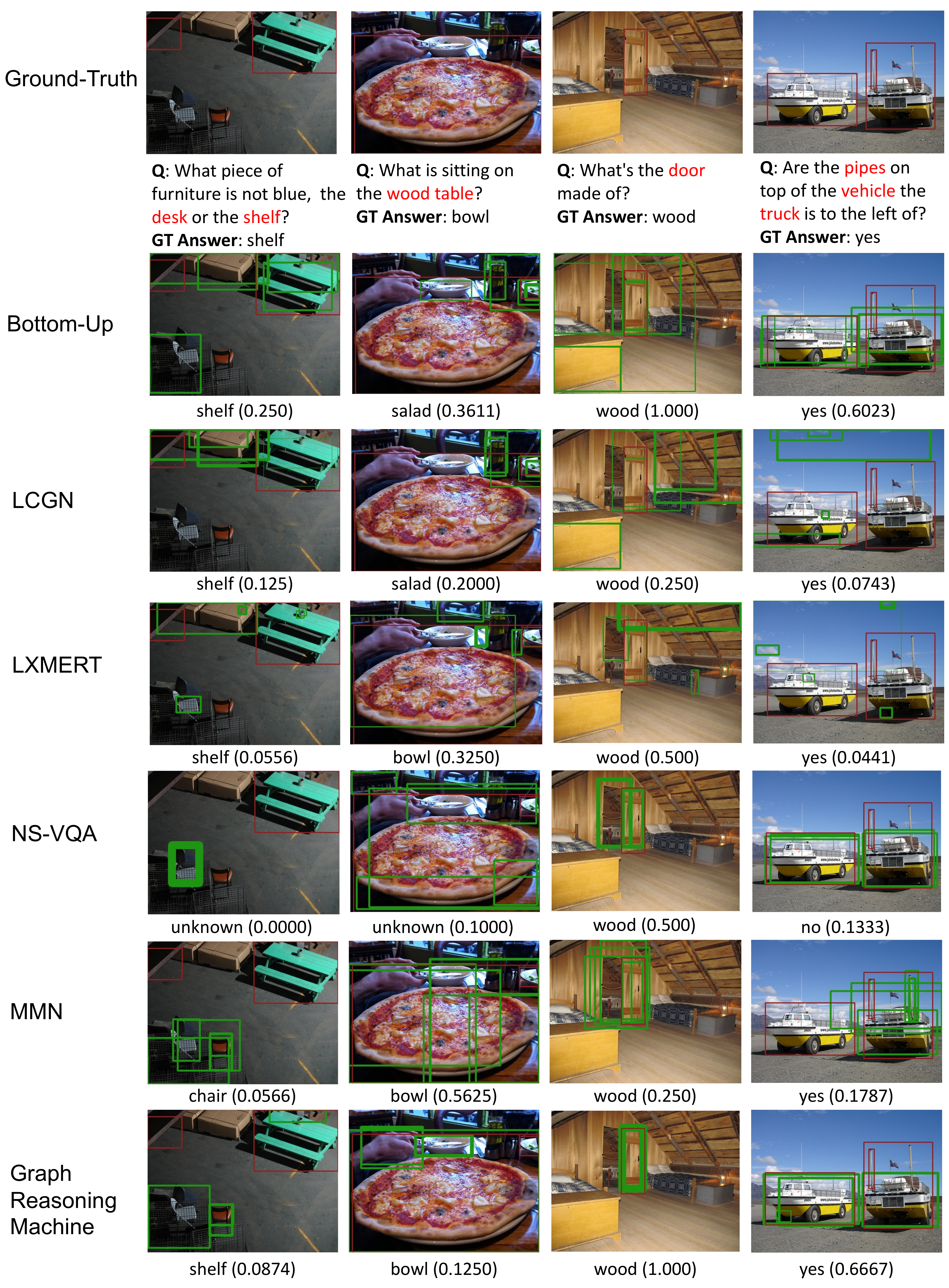}
    \caption{Interpreting grounded regions in different models. From top to bottom, we show ground-truth target objects, grounding results for different models. The red boxes in images represent grounded objects and the green boxes represent attended objects. We show at most five detected bounding boxes from the images for clarity. Below each image, we show the answer to the specific question for each model, with the corresponding grounding scores. The last row is from the proposed Graph Reasoning Machine, which we will introduce in Sec.~\ref{Sec:graph_reasoning_machine}.}
    \label{Figure:Grounding}
\end{figure*}

\xhdr{Overall Accuracy.} Before going into the proposed metrics, we first compare the accuracy of different models on GQA test set in \tbl{Table:AccuracyComparison}. There are a few messages conveyed from these comparisons. First, though using a similar pipeline, LCGN and NSM are more powerful than Bottom-Up, because they use graph neural networks and scene graph representations, respectively. Based on this, they can capture more contextual information from the image to facilitate the visual reasoning. Second, though LXMERT does not use graph representation, it achieves comparable performance to LCGN and NSM. This improvement is arguably brought by leveraging more training data and larger transformer~\cite{vaswani2017attention} to implicitly learning the contextual information. Third, regarding explicit reasoning methods, NS-VQA performs poorly on GQA, because it was originally proposed for a synthetic dataset, and the deterministic symbolic operations in visual reasoning is vulnerable to wrong or incomplete scene graph generations. Similar to NS-VQA, MMN also uses question programs. However, it uses object features instead of symbolic representations. Moreover, it encodes both objects and question programs into hidden feature spaces and then learn the reasoning model end-to-end, which mitigates the aforementioned issue in NS-VQA.

\xhdr{Object Grounding}. In \tbl{Table:Grounding}, we report grounding scores under different subset of questions based on the annotations provided in GQA dataset. We can find MMN outperforms other methods with a significant margin. We suspect this is because it explicitly learns to ground on objects for each reasoning step based on the fine-grained annotations, whereas all other models simply use the final answer as the only supervision for training. Though we expect to learn visual grounding merely using question-answer pairs, these results indicate that object grounding is learnable and the fine-grained grounding annotations in GQA help the model to learn the grounding. Among all models, LXMERT has the lowest grounding scores, even though it uses the GQA validation set during pretraining. Since it is designed to have much more parameters to learn from extra large--scale datasets, it is prone to overfit on some spurious correlations between inputs and answers for a higher accuracy at the cost of grounding scores. Bottom-Up and LCGN have similar grounding performance since they use similar attention scheme to derive the answer. At last, NS-VQA has a poor performance on object grounding as overall accuracy. Based on its transparency with object-centric representation and symbolic programs, we spot that the failure cases mostly come from the defective scene graph generation models on GQA images rather than language understanding from questions and logical reasoning. In Fig.~\ref{Figure:Grounding}, we show the top five referred and grounded objects for different models.

\xhdr{Model Robustness}. We further analyze the models' robustness on the GQA validation set. Here, we ignore the LXMERT model~\cite{tan2019lxmert}, as it included the GQA validation set during pretraining. We show results in Table~\ref{Table:Robustness}. Besides the answer flipping rates, we also report the VQA accuracy before and after the perturbations.

When removing the background objects, we can see the transitions from correct answer to incorrect ones and vice versa for all models. Except for NS-VQA, all the other three models have relatively higher $I\rightarrow C$ flipping rate than $C \rightarrow I$, which thus leads to improved accuracies. This is reasonable since removing the background objects will help the models to focus on the referred and grounded objects. For NS-VQA, however, we see inverse trend. We suspect this is because it uses a purely symbolic scene graph and visual reasoning. 
Removing the background objects will not influence the choice of entities in the remaining scene graph since it relies on pure symbolic matching. Considering its poor grounding performance, the originally correct predictions is easier to become incorrect ones than the other way around. 
In contrast, all other methods use contextualized visual representations through attention mechanisms, we can see more overall interchanges between correct and incorrect answers. 

When foreground objects are removed, all models perform poorly, with a high answer flipping ratio for C$\rightarrow$I and a lower one for I$\rightarrow$C. However, the accuracies remain non-zero due to two potential reasons. First, without foreground objects, models can still leverage spurious correlations between background objects and answers. As we already know, the grounding is not perfect for all models. Second, models may also use language priors to answer questions, as discussed in \cite{goyal2017making}. To validate our suspects, we feed randomized object features to each model. As shown in the right part of Table~\ref{Table:Robustness}, NS-VQA collapses because it uses symbolic reasoning and can hardly leverage the random visual features and language priors. However, all other models can still rely on language information to give an answer. Finally, we show some examples in Fig.~\ref{Fig:Robustness}, where correct answers are flipped to incorrect ones when only one background object is removed. 

\section{Graph Reasoning Machine}
\label{Sec:graph_reasoning_machine}
\begin{figure*}[!ht]
    \centering
    \includegraphics[width=1\linewidth]{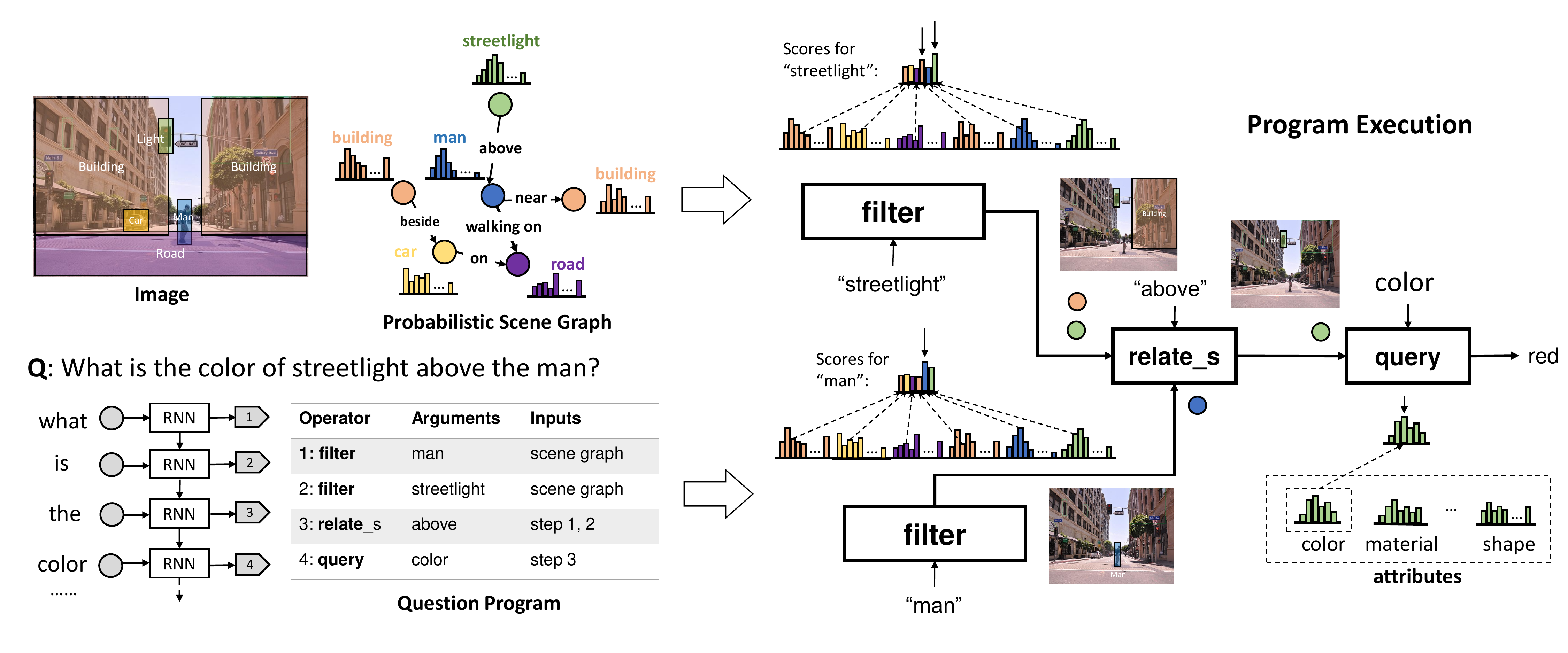}
    \vspace{-18pt}
    \caption{Our graph reasoning machine consists of three components: a) probabilistic scene graph generation; b) question program generation and c) probabilistic program execution.}
    \label{Fig:model}
\end{figure*}

Based on the above diagnosis, we notice that the step-wise reasoning supervisions used in MMN are helpful to achieve better grounding while the symbolic representations used in NS-VQA enable the transparency for diagnosis. Inspired by this, we further propose a diagnostic model called Graph Reasoning Machine while reconciles these two type of models. As shown in Fig.~\ref{Fig:model}, it consists of three components: 1) scene graph generator; 2) program generator and 3) program executor. As we observed earlier, it is hard for the NS-VQA model to handle the realistic images because it is vulnerable to defective visual perception models. 
To remedy these drawbacks, our proposed Graph Reasoning Machine augment the original NS-VQA model by developing probabilistic scene graph representation and neural executor. Based on this new design, we build a teacher-forcing training pipeline to learn the scene graph generator and neural executor jointly. In the following, we will introduce the detailed design choice for the three components in our model.

\subsection{Scene Graph Generator}
\label{Sec:SceneGraph}
In this paper, we carefully design the scene graph representation to facilitate the transparent visual reasoning in that: 1) it should be semantic to maintain its interpretability; 2) it should be compatible with program executor as well as gradient back-propagation. We choose to use the classification score output from the scene graph generator to represent the scene graph. Specifically, a scene graph is denoted by $\bm{G} = <\bm{V},\bm{E}>$ where:
\begin{itemize}
\item $\bm{V}$ is a set of detected object in the image. It contains $K \equiv |\bm{V}|$ objects associated with bounding boxes $\bm{B} \in \mathcal{R}^{K \times 4}$, object classification scores $\bm{S_o} \in \mathcal{R}^{K \times {C}_o}$ and attribute classification scores $\bm{S_a} \in \mathcal{R}^{K \times C_a}$, where $C_o$ and $C_a$ are the number of object and attribute categories, respectively. Both contain the background class.
\item $\bm{E}$ are the edges connecting the objects in the scene graph. Each edge is associated with a predicate classification score. Here, we cover all object pairs in the image to construct a score matrix $\bm{S}_p \in \mathcal{R}^{K \times K \times C_p}$, where $C_p$ is the number of predicate categories including background class.
\end{itemize}

To extract the above scene graph representation, we extend the method proposed in \cite{yang2018graph} and \cite{zellers2018neural} by adding an attribute classification module for each object proposal. Specifically, given the object proposals, we extract the box features $\bm{X}_o = \{\bm{x}_o^i\}$ and union box features $\bm{X}_p = \{\bm{x}_p^{ij}\}$. Then we use three modules $\{\mathrm{\Phi}_{o}, \mathrm{\Phi}_{a}, \mathrm{\Phi}_{p}\}$ to predict $\bm{s}_o$, $\bm{s}_a$ and $\bm{s}_p$, respectively:
\begin{equation}
    s^i_o = \mathrm{\Phi}_{o}(\bm{x}^i_o); \quad s_a^i = \mathrm{\Phi}_{a}(\bm{x}_o^i); \quad {s}_p^{ij} = \mathrm{\Phi}_{p}(\bm{x}_p^{ij})
\end{equation}

In NS-VQA, $\bm{s}_o$, $\bm{s}_a$ and $\bm{s}_p$ are converted to symbolic representations by taking the maximum over the vocabulary, while $\bm{X}_o$ is directly passed to the visual reasoning module in MMN. In our model, however, these classification scores are kept to maintain the semantic information while facilitating the gradient back-propagation to the scene graph generator, as shown in the top-left of Fig.~\ref{Fig:model}.

\textbf{Implementation Details}. To build the scene graph generator, we use ResNeXt-101 \cite{xie2017aggregated} as its backbone and Faster R-CNN \cite{ren2015faster} to detect objects from images. For attribute classification, we reuse the box features from the object detector and add a single-layer MLP to predict the attribute scores. For relationship detection between objects, we use Graph R-CNN \cite{yang2018graph} and freq-prior knowledge \cite{zellers2018neural} to obtain the relationship scores for all object pairs in an image. To ensure the output vocabulary in the scene graph match those in the questions and programs, we collect 1,703 object categories, 618 attribute categories, and 311 relationship categories in the training set. We train the scene graph generator for 150k iterations use stochastic gradient descent (SGD) with learning rate 5e-3, batch size 8, momentum 0.99, and weight decay 5e-4. We reduce the learning rate by factor 10 at iteration 100k and 120k.

\subsection{Program Generator}
In GQA dataset, each question program consists of several operations and each operation consists of three components:
\begin{itemize}
\item \textbf{Operator.} It is the logic skeleton in the program. To cover diverse questions in GQA dataset, we introduce 10 different operators, including \textit{filter}, \textit{query}, \textit{exist}, \textit{verify}, \textit{common}, \textit{relate}, \textit{choose}, \textit{and}, \textit{or}, \textit{not};
\item \textbf{Concept.} This is used to specify the super category for the operator. For different operators, it has different options. For example, for \textit{filter}, the concept can be object concept (\eg, person, animal, furniture), attribute concept (\eg, color, material) or position concept (\eg, vertical position); for \textit{relate}, it can be subject or object;
\item \textbf{Category.} This is the most fine-grained argument in the program, and is also determined by the concept in the operation. For example, if the object concept is \textit{animal}, then the category can be either `none' or a specific category such as dog, sheep, cow, \etc. `none' here means we do not specify any fine-grained category and the operation is applied at the level of concept.
\end{itemize}

Based on the above hierarchical combination, we have in total 7445 different operations. We can build a unique reasoning program by connecting the operations under the dependency constraint for each question. For example in Fig.~\ref{Fig:model}, we can convert question ``what is the color of streetlight above the man?" into a reasoning program consisting of reasoning steps: filter(object(streetlight)), filter(object(man)), relate\_subject(above) and query(color).

Formally, a program generator converts a raw question $\bm{Q} = \{{q}_1, ... , {q}_L\}$ to a reasoning program with $T$ operations $\bm{R} = \{{r}_1, ... , {r}_T\}$. As shown in bottom left of Fig.~\ref{Fig:model}, we use a simple sequence to sequence architecture. Given a question, we first use an embedding layer to encode each word ${q}_i$ to a vector $\bm{x}_q^i$. Then, we feed the sequence of word embeddings into a bidirectional LSTM to obtain the representation for the whole question: $\{\bm{e}_f, \bm{e}_b\} = \mathrm{BiLSTM}(\{\bm{x}_q^{1:L}\})$, where $\bm{e}_f$ and $\bm{e}_b$ are the outputs of LSTM in forward and backward direction, respectively. These two embeddings are concatenated and used as the initial hidden state $\bm{h}^0_r = [\bm{e}_f | \bm{e}_b]$ to another LSTM to derive the reasoning program:
\begin{equation}
    \bm{h}_r^i = \mathrm{LSTM}(\bm{x}_r^i, \bm{h}_r^{i-1}); \quad  y_r^i \sim \mathrm{Softmax}(\bm{W}^T \bm{h}_r^i) \\
\end{equation}
where $\bm{x}^i_r$ is the embedding vector for the input operation token at $i$-th step; $y^i_r$ is the output operation token at $i$-th step which is sampled by sending the hidden state $\bm{h}_r^i$ to a fully-connected layer with parameter $\bm{W}^T$. At training time, $\bm{x}^i_r$ is the embedding for the ground-truth operation token while the predicted operation token from last step at test time. The output dimension of embedding layer for both question and operation tokens are 300. Though simple enough, it turns out that the program generator can achieve almost perfect performance. Given the ground-truth scene graph, we can achieve over 95\% prediction accuracy on the validation set. This decent program generator is the basis for us to perform grounded reasoning.

\textbf{Implementation Details}. We use a sequence-to-sequence model to implement the program generator. On the encoder side, we use a two-layer bi-directional LSTM with hidden size 256. For the decoder, we use a two-layer LSTM with hidden size 512. The vocabulary sizes for encoder and decoder are 2,981 and 7,445, respectively. We train it using Adam optimizer \cite{kingma2014adam} with learning rate 1e-3 for 70k iteration with batch size 256. After we train the program generator, we fix it for the following model training.

\subsection{Neural Program Executor}
Given the probabilistic scene graph and question programs, our reasoning module explicitly executes the reasoning programs to derive the final answer, as shown on the right side of Fig.~\ref{Fig:model}. Since the visual representations are probabilistic, we design a step-wise neural executor which are differential with respect to the scene graph representations.

In the reasoning module, executing a reasoning program $\bm{R} =\{r_1,...,r_T\}$ is a process of traversing on the scene graph $\bm{G}$ along the logic chain until we get the final answer. As we discussed above, we have ten operations in total. We can categorize them into three types. The first type is pointing operations such as \textit{filter}, \textit{relate}, whose inputs and output are both a set of scene graph node indices. The second type is a binary operation whose output is a scalar score indicating a Boolean decision `yes' and `no,' such as \textit{exist} and \textit{verify}. The last one is the answering operation, which outputs the final answer, such as \textit{query}. In the following, we will explain how they are implemented in the neural executor.

\xhdr{Pointing Operation.} Given a set of candidate regions in the scene graph, pointing operation finds the target regions which satisfy the arguments in operation. This involves two operations: \textit{filter} and \textit{relate}. For example in Fig.~\ref{Fig:model}, given an operation $filter(object(streetlight))$, the executor aggregates the classification scores for object category ``streetlight'' from all six object proposals, and then output a subset of two object proposals to the next reasoning step. Formally, for a pointing operation $\bm{r}$, its inputs are $m$ index sets $\bm{Z} = \{\bm{z}_1, ..., \bm{z}_m\}$, where the $i$-th index set $\bm{z}_i=\{z_i^1,...,z_i^{n_i}\}$ is the output from previous operation and contains $n_i$ indices. For ``filter'' operator, $m=1$ means there is one precedent. For ``relate'' operator, there are $m=2$ index sets, and one of them is for subjects while the other one is for objects. As such, it can be further divided into ``relate\_subject'' and ``relate\_object''. Given the probabilistic scene graph $\{\bm{S}_o, \bm{S}_a, \bm{S}_p\}$, pointing operation can be formulated as:
\begin{equation}
    \bm{Z}^{out} = Pointer(\bm{S}_{l}; \bm{Z}^{in}; c)
\end{equation}
where $\bm{Z}^{in}$ and $\bm{Z}^{out}$ are the input and output index set; $l \in \{o, a, p\}$ depends on which concept the arguments specify and $c$ is the category in the corresponding concept.

\xhdr{Query Operation.} Query operation is used to output the final answer to a question. Given the input indices $\bm{Z}^{in}$, the output scores for a query operation is:
\begin{equation}
    \bm{p}^c_l = Query({\bm{S}}_l; \bm{Z}^{in}; c)
    \label{Eq:FinalOperation}
\end{equation}
Passing $\bm{p}^i_l \in \mathcal{R}^{1 \times C_l}$ into a softmax layer, we can obtain the probability distribution over all object or attribute categories and then give the final answer. As shown in Fig.~\ref{Fig:model}, query(color) selects the color prediction scores and then gives the final answer ``red'' with maximal score.

\xhdr{Binary Operation.} Binary operation such as \textit{exist} does not output any target indices. Instead, it outputs a scalar score. For \textit{exist} and \textit{verify} operation, we take the maximum score in the input scores giving $\bm{Z}^{in}$ and $\bm{S}_l$, and then pass it to an one-dimensional linear layer to calibrate its value range. Then we can compare this calibrated scalar value with a pre-determined threshold to determine the boolean output. Depending on the specific reasoning program, binary operation can be either an intermediate operation or the final operation. When it is an intermediate operation, its output score will be sent to the operations merely taking scores as input, such as \textit{and}, \textit{or}, \textit{not}; otherwise, the output scalar score is used to give the Boolean answer `yes' or `no'.

\begin{table*}[t]
\setlength{\tabcolsep}{3.5pt}
\centering
\begin{tabular}{l c c c c c c c c c}
\toprule
\multirow{2}{*}{Model} & Val & \multicolumn{3}{c}{Grounding}  & & \multicolumn{3}{c}{Robustness (w/o BG)}  \\
\cmidrule{3-5}\cmidrule{7-9}
      &    Acc.   &  Correct & Wrong & All & & Acc. & C$\rightarrow$I & I$\rightarrow$C \\
\midrule
 Graph Reasoning Machine    & 58.49  & 61.89 & 41.56 & 52.64 & & 60.78   &  2.83  & 8.24 \\
 -teacher-forcing training & 43.94  &  47.11  & 41.01 & 42.16  & & 40.83 & 4.75 & 3.24  \\
 -probabilistic scene graph & 40.49  & 38.99 & 24.25 & 29.57 & & 38.64 & 5.77 & 1.23 \\
 \midrule
 +ground-truth scene graph  & \textbf{95.26}  &  \textbf{88.18}  & \textbf{50.55} & \textbf{86.51} & & \textbf{94.40} & \textbf{1.23} & \textbf{87.96} \\
\bottomrule
\end{tabular}
\vspace{2pt}
\caption{Ablation study on Graph Reasoning Machine model. ``-teacher-forcing training'' means removing teacher-forcing training; ``-probabilistic scene graph'' means further replace probabilistic scene graph to symbolic ones; ``+ ground-truth scene graph'' means using ground-truth scene graph annotations.}
\label{Table:InDepth}
\end{table*}
\textbf{Teacher-forcing Training}. To leverage the step-wise reasoning annotations in the GQA dataset, we propose a teacher-forcing training strategy to supervise the neural executor. To collect the step-wise supervisions for training the neural executor, we use the ground-truth scene graph $\bm{G}^{gt}$ and run the ground-truth question program $\bm{R}^{gt}$ on it. Through this, we can obtain the selected indices for all pointing operations, boolean values for all binary operations, and the answer for the final operation. We collect the ground-truth indices of detected boxes based on these ground-truth inputs and outputs for each operation. Assume we have the ground-truth input indices $\{\bm{Z}^{gt}\}$ and corresponding boxes $\bm{B}^{gt}$ in the ground-truth scene graph. We first compute the IoUs between detected boxes $\bm{B}$ and $\bm{B}^{gt}$. Then we find the indices of boxes in $\bm{B}$ which have $\text{IoU}>0.5$ with any in $\bm{B}^{gt}$. By this way, we can find all the matched indices in $\bm{B}$ which matches with the ground-truth $\bm{B}^{gt}$ at all reasoning steps.

The step-wise guidance enable us to learn the parameters in the executor like a sequence generation model. We use different losses for different types of operations:
\begin{itemize}
\item \textbf{Pointing operation}. Given the input indices $\bm{Z}^{in}$ and output indices $\bm{Z}^{out}$, we generate binary labels $\bm{y}$ of same size with $\bm{Z}^{in}$ indicating whether the index is contained in $\bm{Z}^{out}$. Then, we pass all the scores specified by the operation concept and category to a sigmoid function and compute the binary cross entropy loss between it and $\bm{y}$. 
\item \textbf{Query operation}. Similar to pointing operation, after we compute the probability $p_l^c$ over all candidate answers, we can compute a cross-entropy loss based on the ground-truth answer.
\item \textbf{Binary operation}. Given the scores from last operation, we compute the output score, and then compute the binary cross entropy loss using the ground-truth Boolean label.
\end{itemize}
After aggregating the losses from different reasoning steps, we then back-propagate the gradients through the probabilistic scene graph to scene graph generator to obtain a better visual representations for the visual reasoning.

\xhdr{Inference.} We run the executor given the predicted scene graph and program. At each pointing operation, we select the top choices in the outputs with scores higher than a threshold 5.0 and then pass them to the next operation. This way, we can obtain a global score in the ending operation, which aggregates the predictions from all previous operations. We empirically find that such a search strategy is slightly better than the greedy counterpart.


\subsection{Ablation Study}
\vspace{2pt}



We use Graph Reasoning Machine as a diagnostic model and perform ablation studies by gradually degrading it to a simple symbolic model. The results are summarized in Table~\ref{Table:InDepth}. Based on the results, we have the following observations.
\xhdr{End-to-end learning with intermediate supervision is helpful}. We first investigate how much benefit the teacher-forcing training can bring to the model. Without teacher forcing training, the neural executor does not need any pre-training on the GQA data. Comparing row 1 and 2 in Table~\ref{Table:InDepth}, we observe a significant drop after removing the teacher-forcing training on all three metrics, accuracy, grounding and robustness. As we discussed above, the scene graph generator have some incorrect predictions. Teacher-forcing training is a way to leverage the step-wise reasoning annotations to adjust the incorrect predictions into correct one by back-propagate the errors to the scene graph generator. These results align with MMN which also exploited the step-wise reasoning supervision. However, the difference is that our Graph Reasoning Machine uses probabilistic scene graph as the visual representations which are more interpretable than the feature embeddings in MMN.

\xhdr{Probabilistic representations outperforms symbolic counterparts}. After removing the teacher-forcing training, we further replace the probabilistic scene graph to a pure symbolic scene graph. Then the visual reasoning becomes a pure symbolic matching, which is exactly the strategy used in NS-VQA. As we can see from row 2 and 3 in Table~\ref{Table:InDepth}, this replacement further hurts the performance across all three metrics. The success of symbolic matching highly depends on the performance of visual representation. A probabilistic scene graph makes the visual reasoning less vulnerable to the incorrect scene graph prediction because it does not require the object/attribute/predicate category of maximal score exactly corresponds to the queried category. For example, to find a car in an image, we only need to find the top object proposals which have relatively higher classification scores for category car, even though all of the objects in the image are not classified as car. Based on this, we can see the superiority of probabilistic scene graph to symbolic one.

\xhdr{Fine-grained visual understanding is critical}. Though we have known from previous works that the visual representation plays an important role in visual reasoning, few papers have investigated how this exactly affects the accuracy and beyond. To do this, we compare a pre-trained scene graph generator and a perfect scene graph provided by GQA. This helps us to understand the main drawbacks in current visual representation that are responsible for the performance gap. We replace the pure symbolic scene graph predicted by our model with the ground-truth one provided by GQA. Compared with the original model (row 1 in Table~\ref{Table:InDepth}), this new setting (row 4) achieves almost perfect performance on validation accuracy. Beyond accuracy, equipped with a perfect vision backbone,the model also achieves much better performance on grounding and robustness. This result indicates that visual backbone plays a critical role for visual reasoning.

Based on the above analysis, we believe a good VQA model requires two essential parts: a good visual understanding and a precise step-wise reasoning. A good visual reasoning model should make a synergy of these two components. For visualization, we show some correct predictions and the corresponding step-wise program executions in top row of Fig.~\ref{Fig:visualizations}. These visualizations can help us to interpret why the model outputs such answers. As we can see, the model can successfully localize the queried objects in the image and further identify their object categories or verify the attribute successfully. For the verify questions, the model needs to localize the objects and then verify whether the target objects have some specific position or share the same color. For the right-most open question sample, it needs first filter the horse and then further relate the localized objects with some specific relationship. Indeed, it is shown that the model can precisely localize the target objects as asked by the question and derive the correct answer. 

\subsection{Failure Analysis}
\begin{figure}[!ht]
    \centering
    \includegraphics[width=0.9\linewidth]{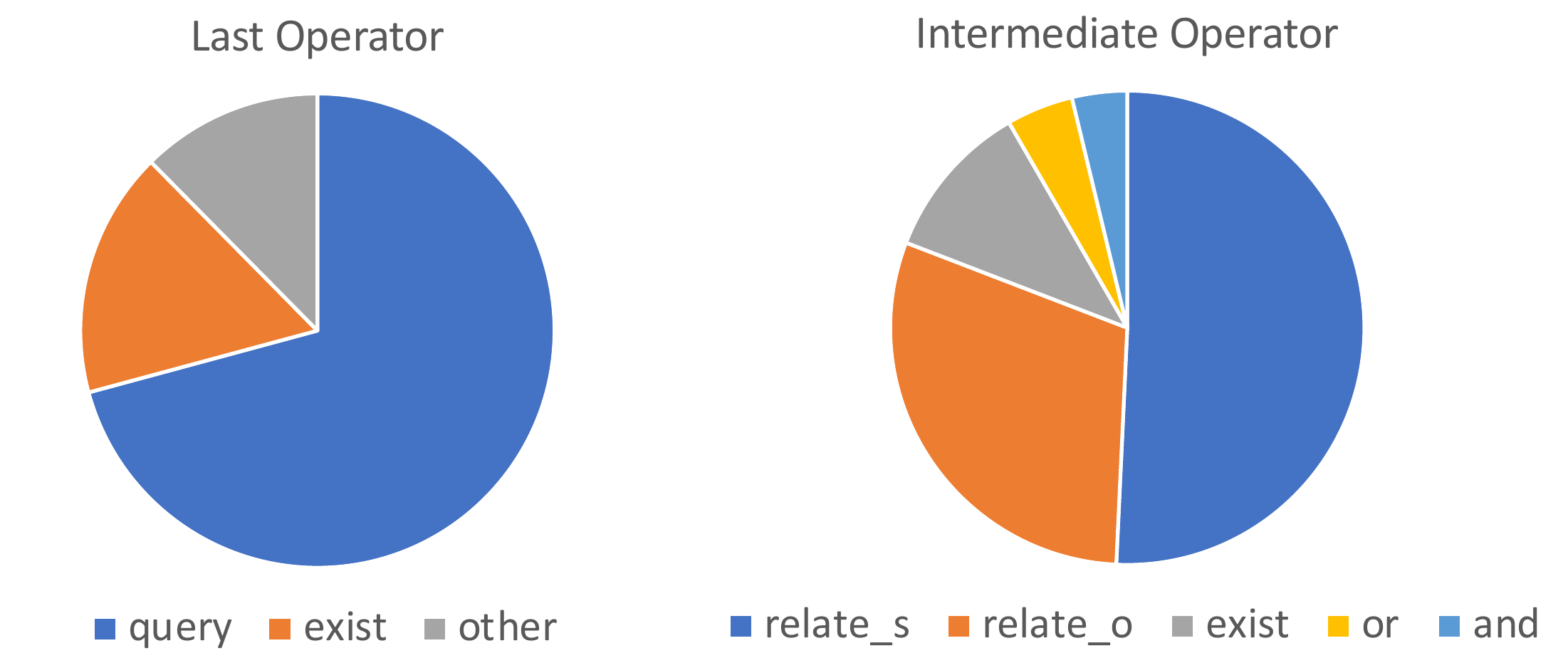}
    \caption{Statistics of failure cases for the augmented NS-VQA model on validation set. Left is the distribution on last operator and right is on intermediate operation.}
    \label{Figure:Execution-error}
\end{figure}
\begin{figure*}[t]
    \centering
    \includegraphics[width=1\linewidth]{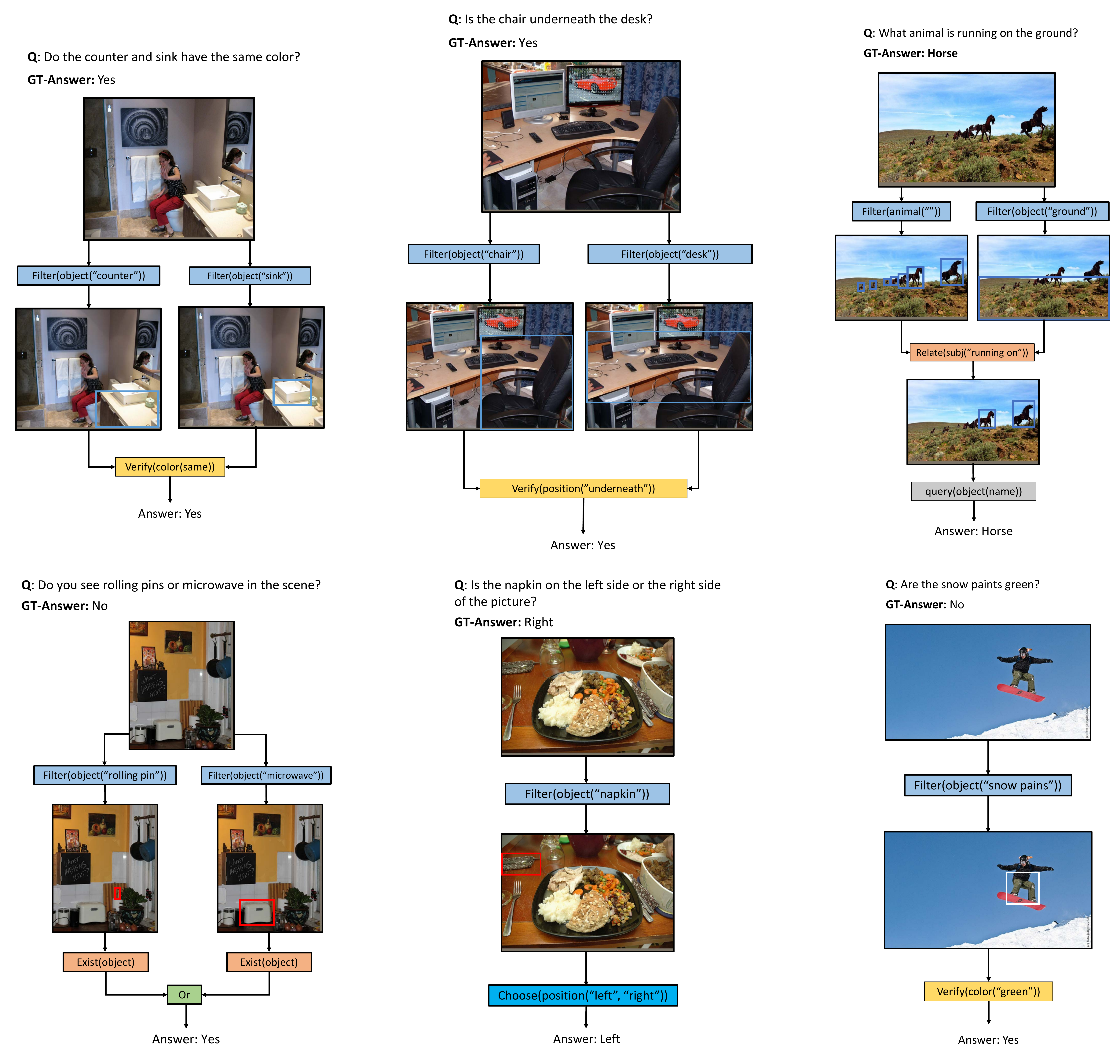}
    \caption{Examples showing step-wise visual reasoning for our graph reasoning machine. The top row shows three samples that our model predicts the correct answers while the bottom row shows three failure cases due to different reasons.}
    \label{Fig:visualizations}
\end{figure*}

Though the Graph Reasoning Machine improves the performance across all metrics compared with the neural-symbolic counterpart, there are still some obvious failures. To get a sense of what the model fails on, we also visualize the reasoning process step-by-step for our graph reasoning machine when it predicts incorrect answer in bottom row of Fig.~\ref{Fig:visualizations}. As we can see in the first sample, the answer becomes ``yes'' because the model has two false-positive localization of ``rolling pins'' and ``microwave''. Though the image does not contain them, the model localize them as labeled by red bounding boxes. For the second sample, there are two napkins on the table. One is on the left while the other one is on the right. The model localizes the left one and thus give the incorrect answer ``left''. For the final sample, the color attribute classification has a wrong prediction which results in an incorrect answer. Taking the merits of transparency in graph reasoning machine, we further get the statistics of failure cases. As shown in Fig.~\ref{Figure:Execution-error} left side, most of the failure cases are from the questions whose last operation is a query and exist. This is expected since the majority of questions are ended with the query and verify operation. On the right side, we further get the statistics on the intermediate operations. Accordingly, more than 75\% failure cases contain \textit{relate} operation. This indicates that the scene graph generator has a clear drawback on relationship detection, though it is one of the state-of-the-art methods in the relationship detection community.

Based on the visualizations and statistics, we summarize the reasons for failures:
\begin{itemize}
\item \textbf{Defective scene graph generator}. We find a primary type of failure cases for our model is caused by defective scene graphs. As mentioned earlier, there are thousands of object categories, and hundreds of attribute and relationship categories. The scene graph generator is far from satisfactory in that its object detection mAP is as low as 4.85, and the attribute recognition and relationship detection also perform very poorly. On the one hand, this poor scene graph representation significantly hinder the model to derive the correct answer. On the other hand, the model can also learn some spurious correlations from the data to predict the answer with a poor visual understanding. As we already see before, we can achieve almost perfect performance using the ground-truth scene graph.
\item \textbf{Multiple referred or grounded objects for a reasoning step}. In the bottom-middle sample of Fig.~\ref{Fig:visualizations}, there are multiple napkins in the image that introduces the ambiguity to the model when answering the positional question. We observe many such ambiguities in the GQA dataset. For example, given a question ``What is in the sky?" for bottom-right sample in Fig.~\ref{Fig:visualizations}, the model can find multiple reasonable answers such as ``man'', ``snowboard'', while only one of them is thought as the ground-truth object since the other objects are not included in the GQA annotations.
\item \textbf{Synonyms and hypernyms in answers}. Though we do not show in this paper, we also observe another type of failure caused by synonyms and hypernyms in answers. Unlike VQA dataset~\cite{antol2015vqa}, where the ground-truth answers are annotated by multiple humans, GQA dataset provides each question a single answer based on the annotations in Visual Genome~\cite{krishna2017visual}. This makes it impossible to measure the correctness as in VQA. The evaluation metric does not accept any answers that are similar to the ground-truth ones. For example, for the question whose ground-truth answer is ``boy'', predicting an answer ``child'' is regarded as totally wrong. Similarly, when the ground-truth answer is ``man'', the answer ``person'' is also thought as wrong. However, we find this is very common in our model and previous models.
\end{itemize}

Based on the above observations, we argue that well-performed vision and reasoning module are both indispensable to obtain a high-accuracy, grounding and robust VQA model. On the other hand, eliminating the ambiguities in both questions and answers in the GQA dataset are also important to establish a better benchmark for evaluating visual reasoning.
\section{Conclusion}

In this paper, we have introduced a set of quantitative metrics for diagnosing visual reasoning models. Leveraging the rich annotations in the GQA dataset, we present object-centric analyses on two important aspects: visual grounding, and robustness. We find that using program-like question representations does improve a model's grounding, but it does not necessarily guarantee its robustness against semantic-level perturbations. We hope our findings inspire future work on designing more accurate, interpretable, and robust models for visual reasoning and broadly machines that may reason over multi-modal data.

\section*{Acknowledgments}
This work was supported in part by the Center for Brains, Minds and Machines (CBMM, NSF STC award CCF-1231216), ONR MURI N00014-16-1-2007, MIT-IBM Watson AI Lab, and MERL.
\ifCLASSOPTIONcaptionsoff
  \newpage
\fi

\bibliographystyle{IEEEtran}
\bibliography{egbib.bib}

\begin{IEEEbiography}[{\includegraphics[width=1in,height=1.25in,clip,keepaspectratio]{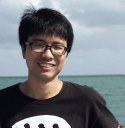}}]{Jianwei Yang}
Jianwei Yang is a senior researcher in Deep Learning Group at Microsoft Research. Before joining Microsoft, he earned his Ph.D. in computer science from School of Interactive Computing at Georgia Institute of Technology, supervised by Prof.~Devi Parikh. His primary research is in the intersection of computer vision, natural language processing (NLP) and embodiment. Specifically, his study focuses on how to leverage structure in image for effective visual understanding, generation and reasoning with natural language.
\end{IEEEbiography}

\begin{IEEEbiography}[{\includegraphics[width=1in,height=1.25in,clip,keepaspectratio]{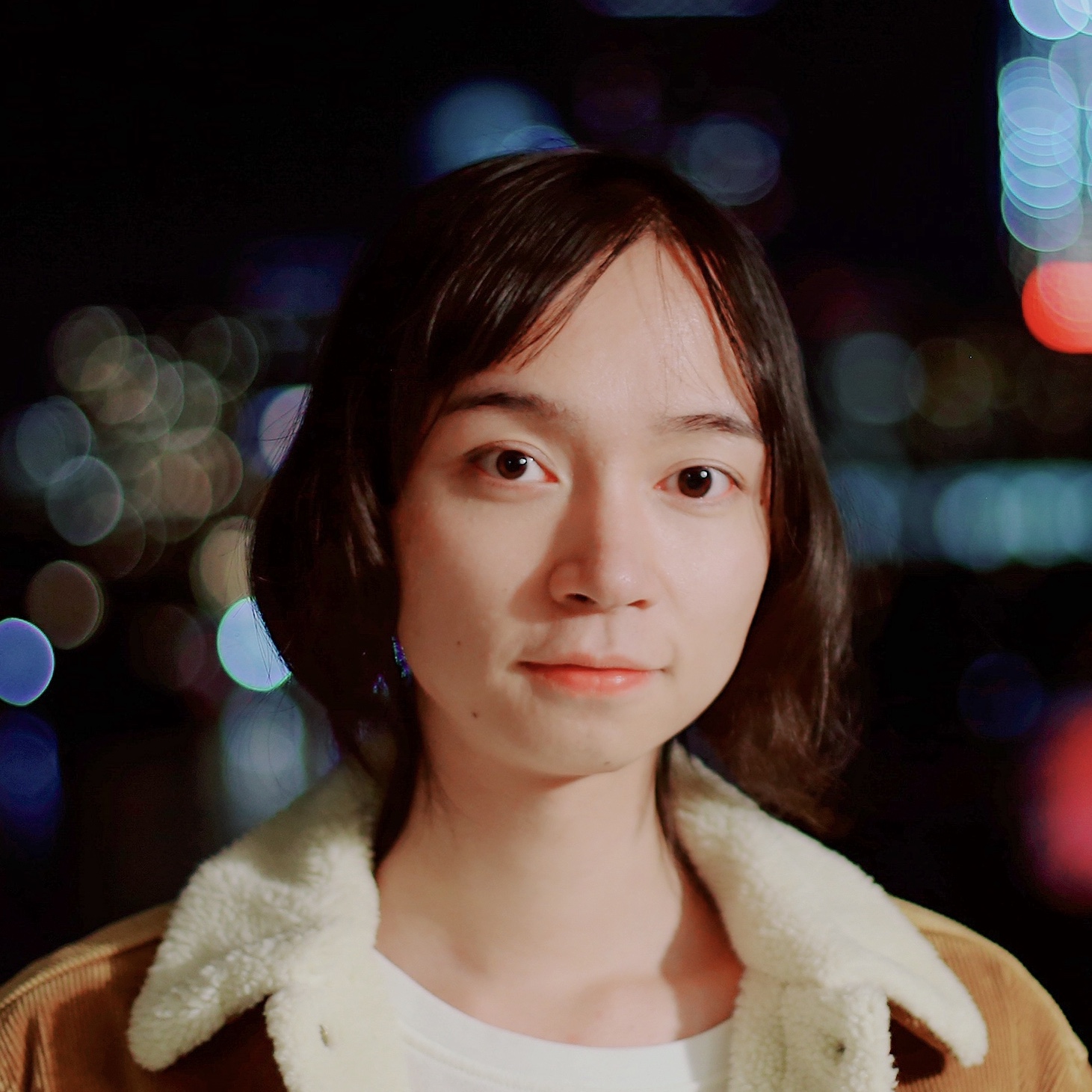}}]{Jiayuan Mao}
Jiayuan Mao is a second-year PhD student in the department of Electrical Engineering and Computer Science at Massachusetts Institute of Technology, advised by Prof. Joshua B. Tenenbuam and Prof. Leslie P. Kaelbling. Mao's research focuses on concept learning, neuro-symbolic reasoning, visual scene understanding, language acquisition, and robotic planning.
\end{IEEEbiography}

\begin{IEEEbiography}[{\includegraphics[width=1in,height=1.25in,clip,keepaspectratio]{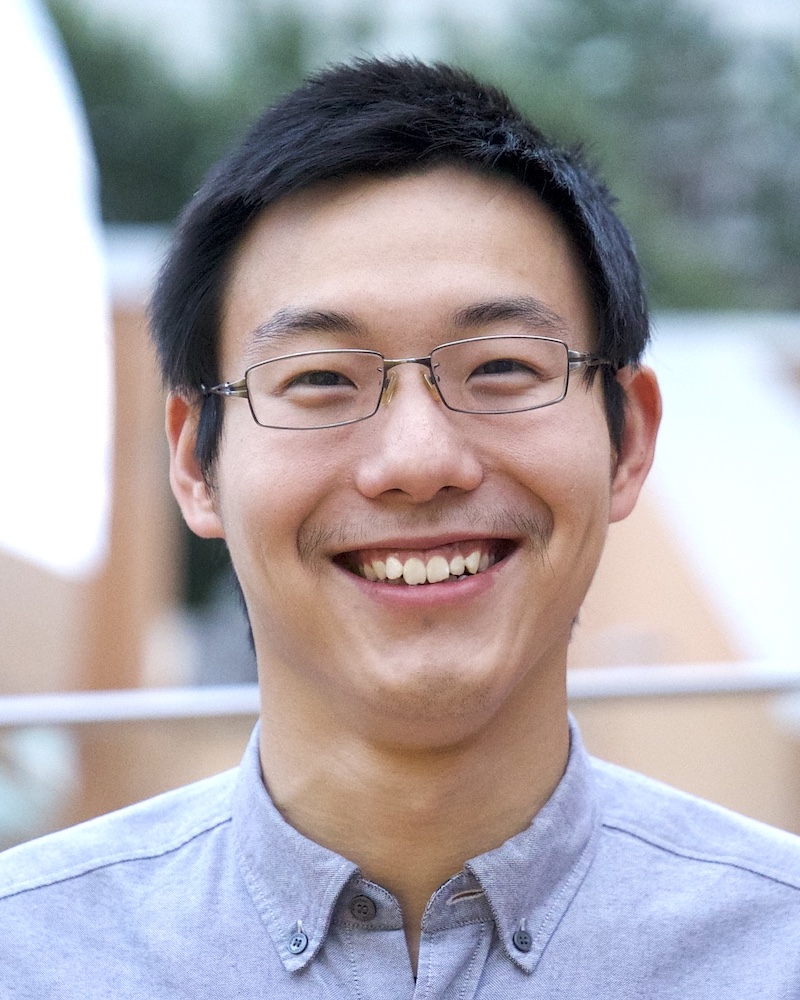}}]{Jiajun Wu} is an Assistant Professor of Computer Science at Stanford University, working on computer vision, machine learning, and computational cognitive science. Before joining Stanford, he was a Visiting Faculty Researcher at Google Research. He received his PhD in Electrical Engineering and Computer Science at Massachusetts Institute of Technology. Wu's research has been recognized through the ACM Doctoral Dissertation Award Honorable Mention, the AAAI/ACM SIGAI Doctoral Dissertation Award, the MIT George M. Sprowls PhD Thesis Award in Artificial Intelligence and Decision-Making, the 2020 Samsung AI Researcher of the Year, the IROS Best Paper Award on Cognitive Robotics, and fellowships from Facebook, Nvidia, Samsung, and Adobe.
\end{IEEEbiography}

\begin{IEEEbiography}[{\includegraphics[width=1in,height=1.25in,clip,keepaspectratio]{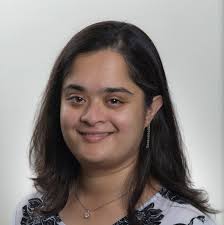}}]{Devi Parikh} is an Associate Professor in the School of Interactive Computing at Georgia Tech, and a Research Scientist at Facebook AI Research (FAIR). Her research interests are in computer vision, natural language processing, embodied AI, human-AI collaboration, and AI for creativity.
She is a recipient of an NSF CAREER award, an IJCAI Computers and Thought award, a Sloan Research Fellowship, an Office of Naval Research (ONR) Young Investigator Program (YIP) award, an Army Research Office (ARO) Young Investigator Program (YIP) award, a Sigma Xi Young Faculty Award at Georgia Tech, an Allen Distinguished Investigator Award in Artificial Intelligence from the Paul G. Allen Family Foundation, four Google Faculty Research Awards, an Amazon Academic Research Award, a Lockheed Martin Inspirational Young Faculty Award at Georgia Tech, an Outstanding New Assistant Professor award from the College of Engineering at Virginia Tech, a Rowan University Medal of Excellence for Alumni Achievement, Rowan University’s 40 under 40 recognition, a Forbes’ list of 20 “Incredible Women Advancing A.I. Research” recognition, and a Marr Best Paper Prize awarded at the International Conference on Computer Vision (ICCV).
\end{IEEEbiography}

\begin{IEEEbiography}[{\includegraphics[width=1in,height=1.25in,clip,keepaspectratio]{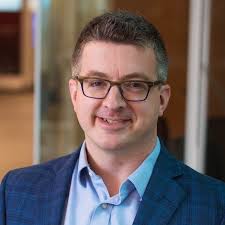}}]{David D. Cox}
 is the IBM Director of the MIT-IBM Watson AI Lab, a first of its kind industry-academic collaboration between IBM and MIT, focused on fundamental research in artificial intelligence. Prior to joining IBM, David was the John L. Loeb Associate Professor of the Natural Sciences and of Engineering and Applied Sciences at Harvard University, where he held appointments in Computer Science, the Department of Molecular and Cellular Biology and the Center for Brain Science. David’s ongoing research is primarily focused on bringing insights from neuroscience into AI research. His past work has spanned a variety of disciplines, from neuroscience experiments in living brains, to the development of machine learning and artificial intelligence methods, to applied machine learning and high performance computing methods.

\end{IEEEbiography}

\begin{IEEEbiography}[{\includegraphics[width=1in,height=1.25in,clip,keepaspectratio]{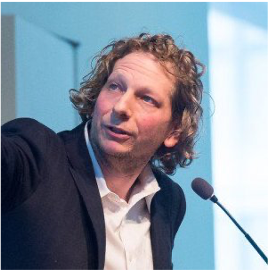}}]{Joshua B. Tenenbaum} is Professor of Computational Cognitive Science at MIT in the Department of Brain and Cognitive Sciences, the Computer Science and Artificial Intelligence Laboratory (CSAIL) and the Center for Brains, Minds and Machines (CBMM). He received his PhD from MIT in 1999, and taught at Stanford from 1999 to 2002. His long-term goal is to reverse-engineer intelligence in the human mind and brain, and use these insights to engineer more human-like machine intelligence. His current research focuses on the development of common sense in children and machines, the neural basis of common sense, and models of learning as Bayesian program synthesis. His work has been published in Science, Nature, PNAS, and many other leading journals, and recognized with awards at conferences in Cognitive Science, Computer Vision, Neural Information Processing Systems, Reinforcement Learning and Decision Making, and Robotics. He is the recipient of the Distinguished Scientific Award for Early Career Contributions in Psychology from the American Psychological Association (2008), the Troland Research Award from the National Academy of Sciences (2011), the Howard Crosby Warren Medal from the Society of Experimental Psychologists (2016), the R\&D Magazine Innovator of the Year award (2018), and a MacArthur Fellowship (2019). He is a fellow of the Cognitive Science Society, the Society for Experimental Psychologists, and a member of the American Academy of Arts and Sciences.
\end{IEEEbiography}



\begin{IEEEbiography}[{\includegraphics[width=1in,height=1.25in,clip,keepaspectratio]{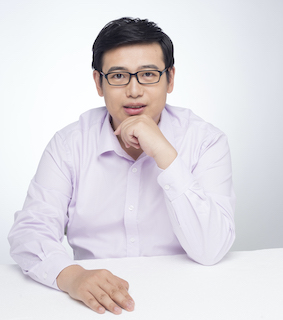}}]{Chuang Gan}
is a principal research staff member at MIT-IBM Watson AI Lab. He is also a visiting research scientist at MIT. Before that, he completed his Ph.D. with the highest honor at Tsinghua University, supervised by Prof. Andrew Chi-Chih Yao. His primary focuses are on machine learning and its applications to computer vision, natural language processing, and robotics. His research works have been recognized by Microsoft Fellowship, Baidu Fellowship, and media coverage from CNN, BBC, The New York Times, WIRED, Forbes, and MIT Tech Review. He has served as an area chair for ICLR, ICCV, AAAI, and IJCAI, and an associate editor for IEEE Transactions on Image Processing, ACM Transactions on Multimedia, IEEE Transactions on Circuits and Systems for Video Technology, and Pattern Recognition. 
\end{IEEEbiography}




\end{document}